\documentclass[lettersize,journal,review]{IEEEtran}
\usepackage{amsmath,amsfonts}
\usepackage{array}
\usepackage{textcomp}
\usepackage{stfloats}
\usepackage{url}
\usepackage{verbatim}
\usepackage{graphicx}
\usepackage{cite}
\usepackage[numbers]{natbib}
\usepackage{latexsym}
\usepackage{amsmath}
\usepackage[T1]{fontenc}
\usepackage[utf8]{inputenc}
\usepackage[ruled,vlined]{algorithm2e}
\usepackage{microtype}
\usepackage{arydshln}
\usepackage{wrapfig,lipsum}
\usepackage{arydshln}
\usepackage{multirow} 
\usepackage{subfigure}
\usepackage{stackengine}
\usepackage{graphicx}
\usepackage{caption}
\usepackage{balance}
\usepackage{xcolor}

\newcommand\xrowht[2][0]{\addstackgap[.5\dimexpr#2\relax]{\vphantom{#1}}}

\SetCommentSty{mycommfont}

\SetKwInput{KwInput}{Input}                % Set the Input
\SetKwInput{KwOutput}{Output}              % set the Output

\begin{document}
\title{Actively Discovering New Slots for Task-oriented Conversation}
\author{Yuxia Wu$^\ast$,
	Tianhao Dai$^\ast$,
	Zhedong Zheng,
	Lizi Liao$^\dag$

	%\thanks{This work was supported in part by the .}.
	\thanks{Yuxia Wu and Lizi Liao are with the Singapore Management University (e-mail: yieshah2017@gmail.com, lzliao@smu.edu.sg)} 
	\thanks{Tianhao Dai is with Wuhan University (e-mail: 	tianhao.dai@outlook.com)} 
	\thanks{Zhedong Zheng is with the National University of Singapore (e-mail: zdzheng@nus.edu.sg)} 
 	\thanks{$^\ast$ Co-first authors with equal contribution.}	
 	\thanks{$^\dag$ Corresponding author.}

}

\markboth{Journal of \LaTeX\ Class Files,~Vol.~14, No.~8, August~2021}%
{Shell \MakeLowercase{\textit{et al.}}: A Sample Article Using IEEEtran.cls for IEEE Journals}

%\IEEEpubid{0000--0000/00\$00.00~\copyright~2023 IEEE}

\maketitle

\begin{abstract}
Existing task-oriented conversational search systems heavily rely on domain ontologies with pre-defined slots and candidate value sets. In practical applications, these prerequisites are hard to meet, due to the emerging new user requirements and ever-changing scenarios.
To mitigate these issues for better interaction performance, there are efforts working towards detecting out-of-vocabulary values or discovering new slots under unsupervised or semi-supervised learning paradigm. However, overemphasizing on the conversation data patterns alone induces these methods to yield noisy and arbitrary slot results. 
%In order to improve the pragmatic utility, a stringent amount of human labelling quota is often available, which offers an authoritative way to obtain accurate and meaningful slot assignments while also brings forward the high requirement of utilizing such quota efficiently.
To facilitate the pragmatic utility, real-world systems tend to provide a stringent amount of human labelling quota, which offers an authoritative way to obtain accurate and meaningful slot assignments. Nonetheless, it also brings forward the high requirement of utilizing such quota efficiently.
Hence, we formulate a general new slot discovery task in an information extraction fashion and incorporate it into an active learning framework to realize human-in-the-loop learning. Specifically, we leverage existing language tools to extract value candidates where the corresponding labels are further leveraged as weak supervision signals. 
Based on these, we propose a bi-criteria selection scheme which incorporates two major strategies, namely, uncertainty-based sampling and diversity-based sampling to efficiently identify terms of interest. We conduct extensive experiments on several public datasets and compare with a bunch of competitive baselines to demonstrate the effectiveness of our method. We have made the code and data used in this paper publicly available\footnote{https://github.com/newslotdetection/newslotdetection}.
\end{abstract}


\section{Introduction}
\IEEEPARstart{W}{ith} the development of smart assistants (\textit{e.g.}, Alexa, Siri), conversational systems play an increasing role in helping users with tasks, such as searching for restaurants, hotels, or general information. Slot filling has been the main technique for understanding user queries in deployed
systems, which heavily relies on pre-defined ontologies \cite{multiwoz,chen2019bert,louvan2020recent,balaraman2021domain,jiao2022enhanced}.  
However, many new places, concepts or even application scenarios are springing up constantly.
Existing ontologies inevitably fall short of hands, which hurts the system performance and reliability. 
As one of the foundation blocks in ontology learning, new slot discovery is particularly crucial in those deployed systems. It not only discovers potential new concepts for later stage ontology construction or update, but also helps to avoid incorrect answers or abnormal actions.

%Therefore, constructing or updating dialogue ontology is a critical but largely ignored task. 
\begin{figure}
	\centering
	%\vspace{+0.4cm}
	\includegraphics[scale=0.49]{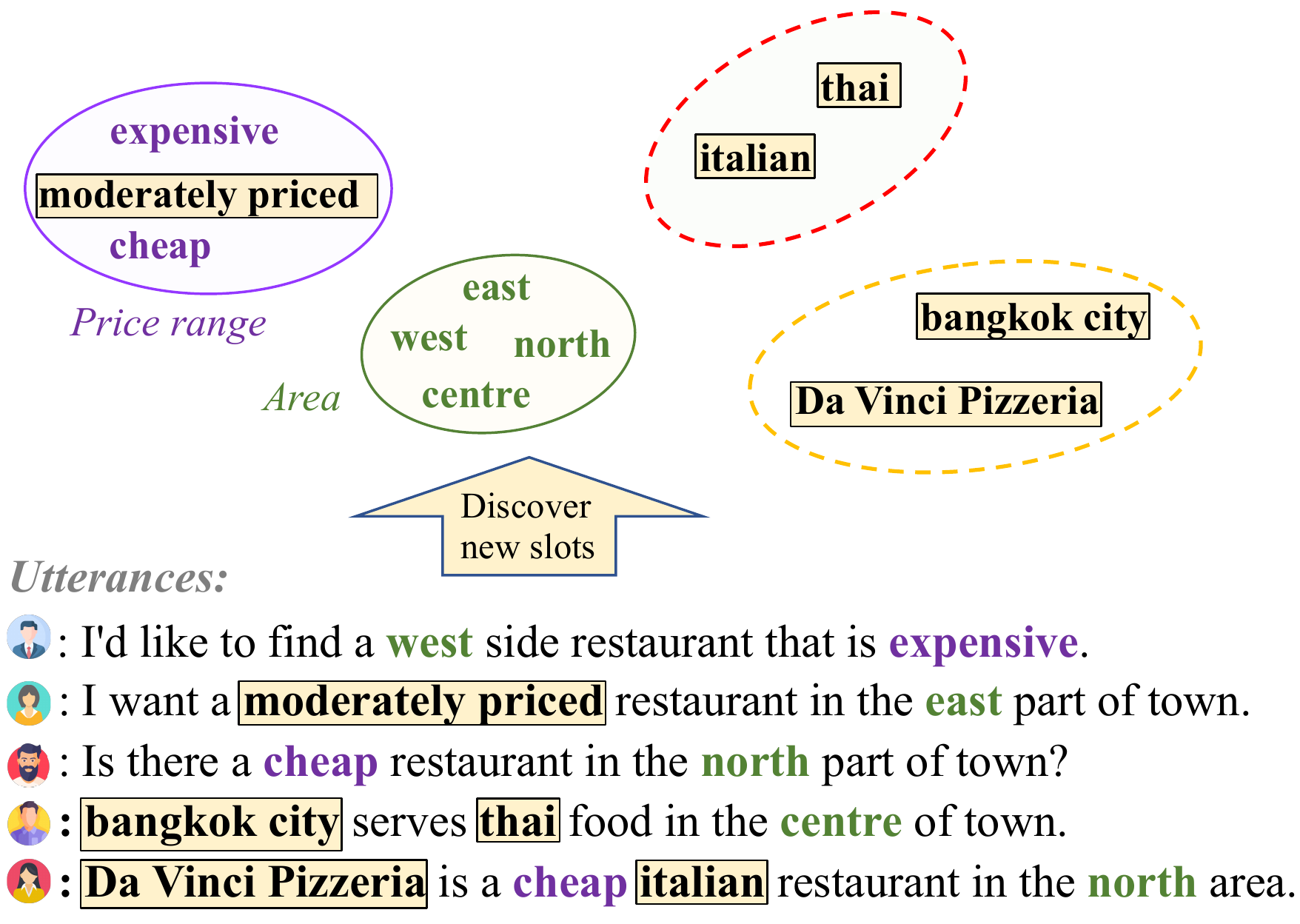}
	%\vspace{-0.4cm}
	\caption{Illustration of the general new slot discovery task. It not only finds new values for predefined slots (e.g., \textit{Price range}, \textit{Area} in solid circles), but also discovers new slots with corresponding values (in dotted circles). Those in bold font are extracted value candidates.}
	\label{example}
 \vspace{-0.4cm}
\end{figure}

Generally speaking, new slot discovery requires handling two situations properly as illustrated in Figure \ref{example}: to recognize out-of-vocabulary values for pre-defined slots, and to 
group certain related values into new slots (as in dotted circles). 
Existing works tend to \textbf{separate} these two situations into two independent tasks for ease of modeling:
(1) In the first new value discovery task, several pioneering works leverage
character embeddings to handle the unseen words during training \cite{liang2017combining} while others harness the copy mechanism for selection \cite{zhao2018improving} or leverage BERT \cite{chen2019bert} for value span prediction.
There are also methods making use of background knowledge \cite{he2020learning, DBLP:conf/acl/CoopeFGVH20}. 
The core of such methods lies in finding the patterns or relations of existing values among predefined slots.
(2) For the second new slot scenario, it is more complicated and requires grouping the values into different slot types even without knowing the exact number of new slots. To simplify the problem, \citet{wu2021novel} proposed a novel slot detection task  without differentiating the exact new slot names. For a more realistic setting, other researchers adapt transfer learning to leverage the knowledge in the source domain to discover new slots in the target domain \cite{shah2019robust,wang2021bridge}. They assume that the slot descriptions or even some example values are available. However, such availability is still less likely in practice.
Hence, another line of research efforts seek help from existing tools such as semantic parser or other information extraction tools to gain knowledge \cite{chen2013unsupervised,chen2014leveraging,zeng2021automatic,siddique2021linguistically}. Nonetheless, such methods suffer from the noisy nature of dialogue data and require intensive human decisions in various processing stages and settings. %Besides, these works mainly focus on sequence labeling methods based on RNN or BERT model, which overemphasize the sequential dependency among different slot labels. 

The current popular sequence labeling way emphasizes the relationship patterns in word or token sequences and labels, which is less sufficient for out-of-scope slots. As the new value and new slot discovery are inherently intertwined, we propose to adopt an Information Extraction (IE) fashion to tackle them concurrently as a general new slot discovery task. Candidate values are extracted firstly, which are then leveraged to find group structures. Nonetheless, if we obtain group structures purely based on data patterns, the resulting slots will tend to be noisy and arbitrary. 
Fortunately, a stringent amount of human labeling quota is usually available to facilitate the pragmatic utility, which offers an authoritative way to obtain accurate and meaningful slot assignments. To utilize such quota efficiently, a viable way is to adopt the active learning (AL) scheme \cite{freund1997selective, tur2005combining, liu2018learning, hazra2021active2} to progressively select and annotate data to expand our slot set. In general, existing active learning methods can be categorized into two major groups based on the sample selection strategy: uncertainty-based, diversity-based \cite{kim2020deep}. The former tries to find hard examples using heuristics like highest entropy or margin and so on \cite{dasgupta2011two,liang2017combining,siddhant2018deep,zheng2021rectifying}, while the latter aims to select a diverse set to alleviate the redundancy issue  \cite{das2010probabilistic,ash2019deep,gissin2019discriminative}. Although there are works combining these two kinds of strategies and working well on the sequence tagging task \cite{liu2018learning, shen2018deep, hazra2021active2}, their success is not directly applicable to our setting, because one sequence might contain multiple different slots and the goal of finding new slots is less emphasized in these sequence labeling models when label sets are known.

In this work, we formulate the general new slot discovery task in an information extraction fashion and design a Bi-criteria active learning scheme to efficiently leverage limited human labeling quota for discovering high-quality slot labels. The IE task 
can naturally fit the proposed active learning procedure. It allows our method to focus on only one of the slots in the input sentence during the sample selection.
Specifically, we make use of the existing well-trained language tools to extract value candidates and corresponding weak labels. Being applied as weak supervision signals, these weak labels are integrated into a BERT-based slot classification model via multi-task learning to guide the training process. With the properly trained model, we further design a Bi-criteria sample selection scheme to efficiently select samples of interest and solicit human labels. In particular, it incorporates both uncertainty-based sampling and diversity-based sampling strategies via maximal marginal relevance calculation, which strives to reduce redundancy while maintaining uncertainty levels in selecting samples.
 
To sum up, our contributions are three-fold: 
\begin{itemize}
	\item We formulate a general new slot discovery task that wrap up the new value and new slot scenarios. Formatted in an IE fashion, it benefits from existing language tools as weak supervision signals.
	\item We propose an efficient Bi-criteria active learning scheme to identify new slots. In particular, it incorporates both uncertainty and diversity-based strategies via maximal marginal relevance calculation.
	\item Extensive experiments verify the effectiveness of the proposed method and show that it can largely reduce human labeling efforts while maintaining competitive performance.
\end{itemize}
\section{Related Work}
\subsection{Out-of-Vocabulary Detection}
New slot discovery aims to discover potential new slots for conversation ontology construction or update. It is closely related to the Out-of-Vocabulary (OOV) detection task that aims to find new slot values for existing slots. Under this task setting, the slot structures are predefined. For example, given the slots such as \textit{Price range} and \textit{Area}, it aims to find new values such as \textit{moderately priced} to enrich the value set. \citet{liang2017combining} combined the word-level and character-level
representations to deal with the out-of-vocabulary words. They treated the characters as atomic units which can learn the representations of new words. 
\citet{zhao2018improving} leveraged the copy mechanism based on pointer network. The model is learned to decide whether to copy candidate words from the input utterance or generate a word from the vocabulary. \citet{chen2019bert} trained BERT \cite{kenton2019bert} for slot value span prediction which is also capable of detecting out-of-vocabulary values. \citet{he2020learning} proposed a 
background knowledge enhanced model to deal with OOV tokens. The knowledge graph provides explicit lexical relations among slots and values to help recognize the unseen values. More recently, \citet{DBLP:conf/acl/CoopeFGVH20} regarded the slot filling task as span extraction problem. They integrate the large-scale pre-trained conversational model to few-shot slot filling which can also handle the OOV values.
%\citet{chen2019transfer} proposed a transfer learning method for sequence labeling. It can transfer the knowledge from the source domain to target domain which contains new categories. %\cite{shah2019robust} focused on zero-shot cross-domain slot filling task and    

%there are various methods being applied such as transfer learning \citet{chen2019transfer}, few/zero-shot learning \cite{shah2019robust} and making use of background knowledge \cite{yang2017leveraging}, \textit{etc}.
%The core of such methods lies in finding and leveraging the patterns of existing values or various relations among these predefined slots. In our work, the novel term detection task does not over-emphasize on discriminating whether a new term belongs to existing slots or not. We do not rely on a predefined complete slot structure. Instead, we assume that it might be easier to decide the structure among terms when the term candidate pool is ready.

\subsection{New Slot Discovery}
Finding new slots requires proper estimation of the number and structural composition of new slots. As this is hard, there are efforts assuming that the slot descriptions of new slots or even some example values for these slots are available. These slot description or example values are directly interacted with user utterances to extract the values for each new slot individually \cite{bapna2017towards,shah2019robust,lee2019zero,hou2020few,oguz2021few}. However, the over-reliance on slot descriptions hinders the generality and applicability of such methods. There are works trying to ignore such information. For example, \citet{wu2021novel} proposed a novel slot detection task to identify whether a slot is new or old without further grouping them into different classes. The Out-of-Distribution detection algorithms (such as MSP \cite{hendrycks2016baseline} and GDA \cite{xu2020deep}) are leveraged to fulfill the task. However, they only worked on simulated datasets and the task scenario is oversimplified.

Hence, researchers proposed a two-stage pipeline which first extracts slot candidates and values using a semantic parser or other information extraction tools, and then utilizes various ranking or clustering methods to pick out salient slots and corresponding values. For example, \citet{chen2013unsupervised} combined semantic frame parsing with word embeddings for slot induction. In the same line, \citet{chen2014leveraging} further constructed lexical knowledge graphs and performed a random walk to get slots. Although the language tools provide useful clues for the later stage slot discovery, such methods suffer from the noisy nature of dialogue data and the selection, ranking process requires intensive human involvement. To mitigate this issues, \citet{hudevcek2021discovering} extended the ranking into an iterative process and built a slot tagger based on sequence labeling model for achieving higher recall. Nonetheless, the model still relies on obtained slots in the former iterative process which requires intensive human decisions. 

%There are also semi-supervised methods for discovering novel intents \cite{lin2020discovering, zhang2021discovering}. The general idea is learning the prior knowledge from the labeled data to generate pseudo labels for training. These kinds of methods can be adapted to the new slot discovery task. However, although they can utilize the dataset effectively, the pseudo labels will bring some noise to the model not well trained which leads to inaccurate performance. More recently, \citet{wu2021novel} propose a novel slot detection task identifying whether a slot is new or old without further grouping them into different classes. They only work on simulated datasets and directly apply the Out-of-Distribution detection algorithms (such as MSP \cite{hendrycks2016baseline} and GDA\cite{xu2020deep}) to fulfill the task. 
%\vspace{-0.2cm}
\subsection{Active Learning}
Deep neural networks have recently produced state-of-the-art results on a variety of supervised learning tasks. Nonetheless, many of these achievements have been limited to domains where large amounts of labeled data are available.
Active learning (AL) \cite{freund1997selective} reduces the need for large quantities of labeled data by intelligently selecting unlabeled examples for expert annotation in an iterative process \cite{mccallumzy1998employing,yu2017active}. 
Recently, AL in conjunction with deep learning has received much attention. Several studies have investigated active learning (AL) for natural language processing tasks to alleviate data dependency \cite{shen2018deep}. 
There are two major sample selection strategies for active learning, namely, uncertainty-based and diversity-based sampling \cite{ash2019deep}. Uncertainty-based sampling selects new samples that maximally reduce the uncertainty the algorithm has on the target classifier. In the context of linear classification, \citet{schohn2000less,tur2005combining} proposed such methods that query examples that lie closest to the current decision boundary. Some other approaches have theoretical guarantees on statistical consistency \cite{hanneke2014theory,balcan2006agnostic}. These methods have also been recently generalized to deep learning, e.g., \citet{siddhant2018deep} experimented with Bayesian uncertainty estimates beyond the least confident standards explored by former works. However, a previous work points out that focusing only on the uncertainty leads to a sampling bias \cite{dasgupta2011two}. It creates a pathological scenario where selected samples are highly similar to each other. This may cause problems, especially in the case of noisy and redundant real-world datasets. 

Another approach is diversity-based sampling, wherein the model selects a diverse set such that it represents the input space without adding considerable redundancy \cite{sener2018active}. Certain recent studies for classification tasks adapt the algorithm BADGE \cite{ash2019deep}. It first computes embedding for each unlabeled sample based on induced gradients, and then geometrically picks the instances from the space to ensure their diversity. Inspired by generative adversarial learning, \citet{gissin2019discriminative} selected samples that are maximally indistinguishable from the pool of unlabeled examples. 

More recently, several existing approaches support a hybrid of uncertainty-based sampling and diversity-based sampling \cite{kim2020deep}. For instance, \citet{hazra2021active2} proposed to leverage sample similarities to reduce redundancy on top of various uncertainty-based strategies as a two-stage process. Better performances achieved signal a potential direction to further reduce human labeling efforts. At the same time,
%More recently, leveraging transfer learning via deep pre-trained models enables the selection model to yield remarkable performance when only hundreds of labeled training instances are available. For 
\citet{shelmanov2021active} investigated various pre-trained models and applied Bayesian active learning to sequence tagging tasks. Experiments also showed better performance as compared to those single strategy based ones.
In our work, we take advantage of pre-trained models such as BERT, and design a Bi-criteria active learning scheme to possess the benefits of both uncertainty-based and diversity-based sampling strategy.

The main differences between the proposed method and the related work are: 1) Our method only needs a few annotated data rather than extra prior knowledge such as slot descriptions or example values. 2) Compared with the new slot detection method, our model further organizes the new slots into different categories. 3) Compared with the weak supervised or unsupervised methods, our method mitigates human efforts such as selecting and ranking the candidate slots. Besides, we formulate slot discovery as an information extraction task to better capture the relationship among different values.
\section{Problem Formulation}
\subsection{Background}
Current task-oriented dialogue systems heavily rely on slot filling where an ontology $\mathcal{O}$ is usually provided with slots $\mathcal{S}$ and some candidate values. To find values for slots, existing approaches typically model it as a sequence labeling problem using RNN \cite{mesnil2014using,goo2018slot,zhu2020prior} or pre-trained language models such as BERT \cite{liao2021dialogue}. Given an utterance  $X = \{x_1, x_2, \cdots, x_N\}$ with $N$ tokens, the target of slot filling is to predict a label sequence $L = \{l_1, l_2, \cdots, l_N\}$ using BIO format. Each $l_n$ belongs to three types: B-slot\_type, I-slot\_type, and O, where B- and I- represent the beginning and inside of one candidate value, respectively, and O means the token does not belong to any slot.

\subsection{New Slot Discovery in an IE Fashion}
Though popular \cite{wu2021novel,hudevcek2021discovering}, the sequence labeling framework does not naturally fits the new slot discovery task well. First, the label set is not known beforehand in realistic settings. Second, sequence labeling models rely heavily on the linguistic patterns in utterance and the dependencies among the labels in label sequence. In fact, the candidate values are diverse in nature, they may reside in rather different dialogue contexts and show various linguistic patterns. Hence, it might be hard for sequence labeling models to take the dependencies between labels in the sequence into account \cite{huang2015bidirectional,ma2016end}. Last but not the least, one utterance usually contains semantics about multiple slots. In this way, the sample selection step in active learning methods has to consider the scores of all tokens in a sentence, which leads to a mixed measure of the mutual interaction between different slots.  

From another perspective, the general new slot discovery task covers the new value and new slot scenarios, which naturally fits the information extraction framework where we first extract value candidates, then dispatch or group them into different slots. Under this framework, there are many off-the-shelf language tools available to assist the candidate values extraction and provide weak supervision signals to further assist the grouping stage \cite{hudevcek2021discovering}.

%considering that the goal of new slot discovery is to find the candidate values and identify the relationship among the values, we regard the new slot discovery as an information extraction task. Firstly candidate values and weak labels are extracted by the external NLP tools. Then the final slot types are decided via the Bi-criteria active learning module. In what follows, we will introduce the two steps in detail.

%Fortunately, many language tools are available to assist the candidate values extraction and additional slot type identification which further contributes to the new slot discovery task \cite{hudevcek2021discovering}.

\subsubsection{Candidate Value Extraction and Filtering} 
To reduce the labeling effort, we first extract candidate values which can be a single word or a span of words conveying important semantic. Inspired from \cite{hudevcek2021discovering}, we adopt a frame semantic parser SEMAFOR \cite{das2010probabilistic,das2014frame} and named entities recognition (NER) to extract candidate values. Other methods can also be applied in general such as semantic role labeling (SRL) \cite{palmer2010semantic} or keyword extraction \cite{hulth2003improved}. The SEMAFOR is trained on annotated sentences in FrameNet \cite{baker1998berkeley}. By using SEMAFOR in our corpus, we can extract all semantic frame elements and lexical units from the semantic parsing results as candidate values. Here, we utilize a simple union of results provided by all annotation models \footnote{If the same token span is labeled multiple times by different annotation sources, the span is more likely to be considered as a candidate term.}. The tools also provide labels for the candidate values which can be regarded as weak signals for further model design.

Since the semantic tools are trained on a general corpus, there are some irrelevant values for the conversational search scenario. Thus we further conduct value filtering via some simple rules. In detail, we remove the stop words based on the NLTK tool and the words with lower frequency than a predefined threshold. Besides, we also delete these frequently appear but obviously less useful terms such as the words `then', `looks', `please', `know', and so on.  

\subsubsection{Our New Slot Discovery Formulation}
Different from the general setting of slot filling, we assume that the large-scale labeled training data is limited in the new slot discovery scenario. It is a realistic setting for building conversational agents in new domains or new task settings. Therefore our setting is that we have a set of limited labeled data $D_l$ and a large amount of unlabeled data $D_u$ containing new slot types. %This kind of setting is suitable for both semi-supervised methods and active learning methods. As mentioned before, there are some shortages of semi-supervised methods. 
We design an active learning scheme to efficiently make use of limited human labeling resources for accurate new slot discovery. 

Formally, given a candidate value $X^{i+k}_i = \{x_i, \cdots, x_{i+k}\}$ with $k+1$ tokens extracted from the utterance \textit{X}, our goal is to identify the slot type $y$ of $X^{i+k}_i$. 
Although we only have limited labeled data $D_l$ which contains a set of ($X^{i+k}_i$, $X$, $y$) tuples at the beginning, we will iteratively select and annotate a sample set $S$ from $D_u$ to enrich the data $D_l$ in our active learning scheme. Besides, we also have the weak label $y_{weak}$ for the candidate value $X^{i+k}_i$ which provides additional useful semantics for our model training and sample selection.    

Note that the general new slot discovery task not only covers existing ontology update which identifies new candidate values and label them correctly to existing ontology $\mathcal{O}_{old}$, but also includes ontology expansion where new slots are added to $\mathcal{O}_{old}$ to get $\mathcal{O}_{new}$.

\section{Bi-criteria Active Learning Scheme}
\label{AL}
\begin{figure*}
	\centering
	%\vspace{+0.4cm}
	\includegraphics[scale=0.5]{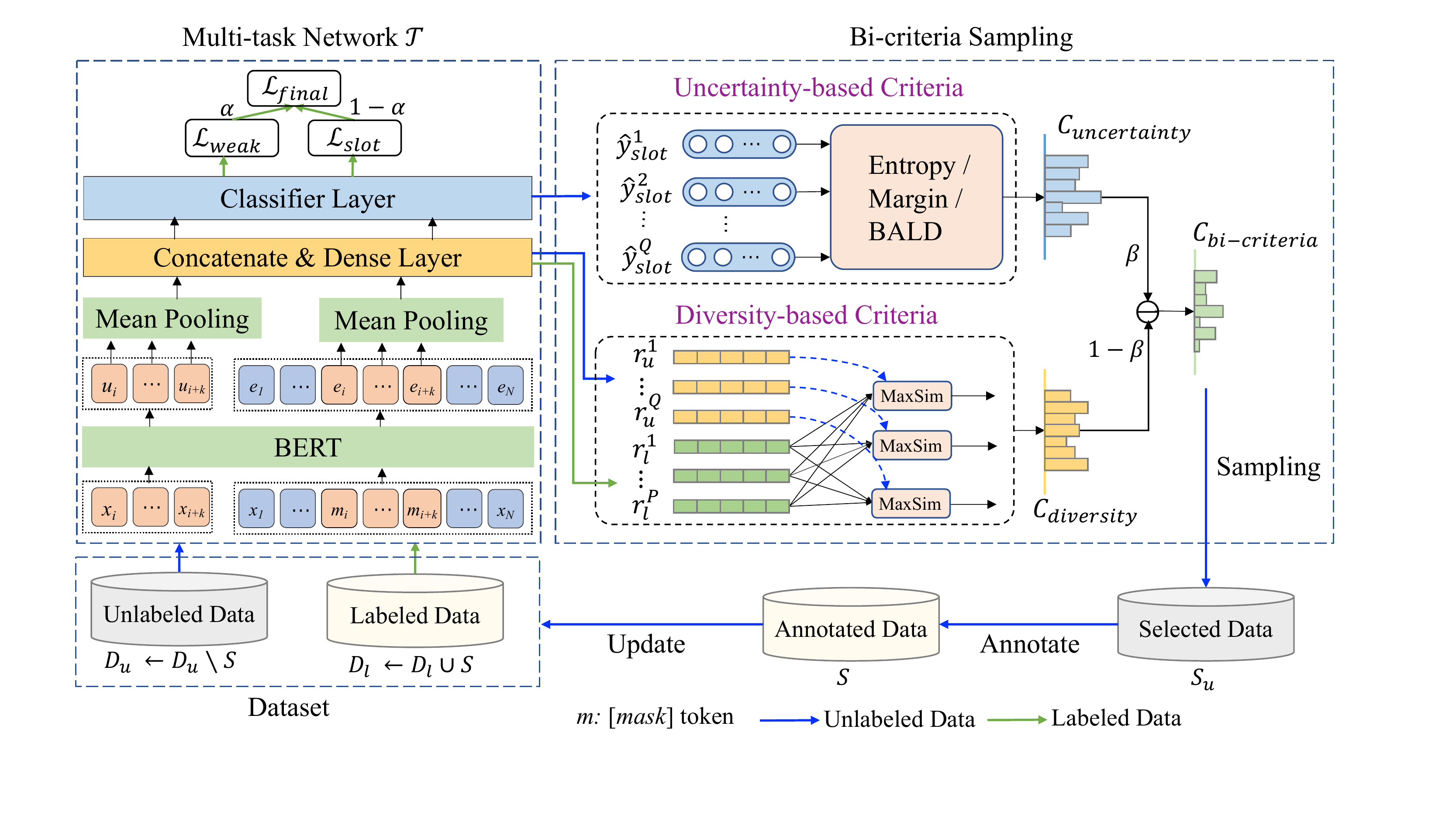}
	%\vspace{-0.4cm}
	\caption{The framework of the proposed Bi-criteria active learning scheme.
		For each iteration, the labeled data is utilized to train the multi-task network $\mathcal{T}$. The unlabeled data is applied to select samples via a Bi-criteria sampling strategy containing both uncertainty and diversity criteria, where BALD is an abbreviation for Bayesian Active Learning by Disagreement. Then the selected samples are annotated and applied to update the dataset for the next loop.}
	%\zznote{ The frame work contains two stages,  Multi-task Training and Bi-criteria Sampling. There is a loop between two stages.  Given the labeled data and unlabeled data, we first conduct Multi-task Training. We extract the feature via pretrained BERT. For the labeled data, we also add some mask to for data augmentation. The outputs are $y_{weak}$ and $y_{slot}$, which are XXXX and XXX respectively. For the second stage, we consider both uncertainty and diversity criteria to sample the potential representative data for experts to annotate. Uncertainty is based on $y$ XXXX, while Diversity is based on XXX $r$. Then we put such annotated data back the dataset and re-train the model. In this way, we could largely save the annotation number, and discover the new concepts efficiently. } }
\label{model}
\end{figure*}

The proposed Bi-criteria active learning method is illustrated in Figure \ref{model}. The dataset contains labeled data and unlabeled data which is updated iteratively via active learning scheme. There are two stages in the iteration loop: multi-task network $\mathcal{T}$ training via labeled data and bi-criteria sampling from unlabeled data. The network $\mathcal{T}$ contains a BERT-based feature extractor and classifier layer. For feature extraction, we concatenate the representations of the candidate values and their context (the \textit{[mask]} token in the position of the values). We train the multi-task network
under the supervision of the weak signals from the NLP tools and the ground truth slot types of the candidate values. 
For the second stage, we first obtain the distributions of classification probabilities and representation features via the trained model $\mathcal{T}$. Then a Bi-criteria strategy is specially designed to incorporate both uncertainty and diversity to select samples. The uncertainty is measured by the characteristics of the probability $\hat{\textbf{y}}_{slot}$ via different strategies. The diversity is computed based on the representations of each sample. Two criteria are integrated by a balanced weight. Finally, the selected samples $\mathcal{S}_u$ are annotated and are applied to update the dataset for the next loop. We will introduce more details about our framework in the following parts.

%The labeled data $D_l$ is used to leaning the multi-task network $\mathcal{T}$. For the unlabeled data $D_u$, we firstly obtain the distributions of classification probabilities and representation features via the trained model $\mathcal{T}$. Then a Bi-criteria sampling strategy is applied to select samples $\mathcal{Su}$. Specially, the Bi-criteria active learning scheme is designed to incorporate both uncertainty and diversity via the maximal marginal relevance (MMR) calculation.

The active learning loop is illustrated in Algorithm \ref{alg} for better understanding. The classification model $\mathcal{T}$ is first trained on 5\% of the whole training dataset denoted as $D_l$. And then different active learning strategies can be applied to select unlabeled samples. After that, we annotate the selected samples $\mathcal{S}$ and add them into the labeled dataset $D_l$. The iteration will stop when $|D_u|=0$ which means there is no more unlabeled data (or stop when model performance no longer increases). 

\begin{algorithm}
	\KwData{$D$: training dataset\\
	              %~~~~~~~~~~~$D_2$: auxiliary classification dataset
	              }
	\KwInput{$D_l \gets$ 5\% of dataset $D$ \\ 
	             ~~~~~~~~~~~~$D_u \gets D - D_l$ \tcp*{unlabeled data}
             } 
	\KwOutput{Well-trained model $\mathcal{T}$ for new slots discovery}
	
	\textbf{Initialization:} $D_l$ \\
	$\mathcal{T} \gets $TRAIN($D_l$)  \tcp*{train with init data}
   % $\mathcal{R} \gets $TRAIN($D_2$)  \tcp*{train with full data}
	
	\tcc{Now starts active learning}
	
	\While{$|D_u|$>0}
	{ 
		\tcc{selection}
		$\mathcal{S}_u \gets Select_{Bi-Criteria}(D_u)$\;
		
		$\mathcal{S} \gets  Annotate(\mathcal{S}_u$)\;
		$D_l \gets D_l \cup \mathcal{S}$ \tcp*{update labeled data}
		$D_u \gets D_u \setminus \mathcal{S}$\tcp*{update unlabeled data}
		%\If{$~~i ~~\% ~~\gamma == 0~~$}{  %\tcp{periodically update $\mathcal{R}$}
		%	$\mathcal{R} \gets $Re-TRAIN($D_2$)
		%}
		$\mathcal{T} \gets $Re-TRAIN($D_l$) 
	}
	\caption{Active Learning Scheme}
	\label{alg}

\end{algorithm}

\vspace{-0.5cm}

\subsection{Multi-task Network $\mathcal{T}$}
We first explain the base classification model $\mathcal{T}$. As mentioned before, we have some limited labeled data with ground truth values extracted from the input utterance and the corresponding slot types. We also obtain the candidate values and their weak labels by language tools. To effectively utilize the weak labels, we introduce a multi-task network to integrate them. Generally speaking, the model contains a feature extractor and a classifier layer with two branches, one for ground truth slot label prediction and the other for weak label prediction. Both branches share the same parameters of the feature extractor and are trained simultaneously. We also try an alternative way of using weak labels where two tasks are conducted chronologically. We show the comparison results and detailed analysis in Section \ref{pretrain}. In the following parts, we introduce the feature extractor, classifier layer, and the loss function of the multi-task network.

\subsubsection{Feature Extractor}
%For the feature extractor, 
Feature extraction for candidate value is the foundation of the subsequent processing for new slot discovery. Both the exact value and its context are essential for a task-oriented conversation system to understand the intents of users. 
Therefore, we integrate the two kinds of representations for each candidate value to facilitate further slot discovery. Specifically, we apply the pre-trained BERT model as the backbone for feature extraction. For the inherent representation, we only consider the token sequence in the candidate value $X^{i+k}_i$. For the context representation, we learn the pure contextual semantics in the input utterance with the candidate value masked to avoid its influence. The detailed process is introduced as follows. 

Given a candidate value $X^{i+k}_i = \{x_i, \cdots, x_{i+k}\}$ with $k+1$ tokens in the utterance \textit{X}, the inherent representation is the mean pooling of all the tokens in $X^{i+k}_i$: \begin{align}
	 \textbf{u}_i, \cdots, \textbf{u}_{i+k}  &= BERT( x_i, \cdots, x_{i+k}),  \\
	\textbf{r}_{inherent} &= mean\_pooling( \textbf{u}_i, \cdots, \textbf{u}_{i+k} ),
\end{align}
where ${\textbf{u}_i}$ represents the embedding of the token ${x_i}$ obtained from the BERT model.

For the context representation of the candidate value, we assume that 
if two values have the same context, they should have similar representations for slot discovery. Therefore we replace the tokens belonging to one value in the original utterance $X$ with a special token \textit{[mask]}. In this way, the utterance is reconstructed as $X'= \{x_1, \cdots, \langle [mask]_i, \cdots, [mask]_{i+k} \rangle, \cdots, x_n\}$\footnote{Special tokens such as \textit{[CLS] in beginning and \textit{[SEP]}} at end are omitted for easy illustration.}. We also adopt the BERT model to obtain the representation of each token in $X'$. With the self-attention mechanism in BERT, the \textit{[mask]} tokens aggregate the contextual semantics of the corresponding values. Hence, we adopt mean pooling on the output of these  \textit{[mask]} tokens to obtain the context representation:
\begin{align}
	 \textbf{e}_1, \cdots, \textbf{e}_i, \cdots, \textbf{e}_{i+k},\cdots, \textbf{e}_n =&~ BERT(X'), \\
	\textbf{r}_{context}=& ~mean\_pooling(\textbf{e}_i, \cdots, \textbf{e}_{i+k}),
\end{align}
where $\langle \textbf{e}_i, \cdots, \textbf{e}_{i+k} \rangle$ denotes the embeddings of the \textit{[mask]} tokens in the last hidden layer of BERT.

We concatenate the inherent and context representation and apply one linear layer followed by tanh activation as the final representation of the candidate value as follows:
\begin{equation}
	\textbf{r} =~tanh({\textbf{W}_1}[\textbf{r}_{inherent};\textbf{r}_{context}]^T+\textbf{b}_1),\label{get_r}
\end{equation}
where $\textbf{W}_1$ and $\textbf{b}_1$ represent the learnable weight matrix and bias.

\subsubsection{Classifier Layer}
As mentioned before, we have two classifiers in the multi-task network. Specifically, we introduce two independent fully-connected layers to map the representation into the probabilities over ground truth slot labels and weak labels given by language tools, \textit{i.e.}: 
\begin{align}
	\hat{\textbf{y}}_{slot}' =&~Softmax({\textbf{W}_2}r^T+\textbf{b}_2),\\
    \hat{\textbf{y}}_{weak} =&~Softmax({\textbf{W}_3}r^T+\textbf{b}_3),
\end{align}
where $\textbf{W}_2$, $\textbf{W}_3$, $\textbf{b}_2$ and $\textbf{b}_3$ represent the learnable weight matrices and biases; $\hat{\textbf{y}}_{slot}'$ and $\hat{\textbf{y}}_{weak}$ represent the predicted probability over all slot labels and weak labels respectively.

\subsubsection{Loss Function}

It is worth noticing that not all the slot labels may have been discovered during training so we apply a label mask to prevent the leakage of unknown labels, which is implemented as:
\begin{align}
	\hat{\textbf{y}}_{slot} =\hat{\textbf{y}}_{slot}' \odot \textbf{m},\label{mask}
\end{align}
where $\textbf{m}$ is a vector with the same dimension as $\hat{\textbf{y}}_{slot}'$ and the $i_{th}$ dimension $m^{(i)}=1$ means slot label $i$ has been known while $m^{(i)}=0$ means unknown; $\odot$ represents element-wise multiplication.

Finally, for each sample, given two predicted probability distributions $\hat{\textbf{y}}_{slot}$ and $\hat{\textbf{y}}_{weak}$, the final loss is constructed as:
\begin{align}
	\mathcal{L}_{final} = (1-\alpha)  \mathcal{L}(\hat{\textbf{y}}_{slot}, \textbf{y}_{slot}) + \alpha \mathcal{L}(\hat{\textbf{y}}_{weak}, \textbf{y}_{weak}),\label{final_loss}
\end{align}
where $\mathcal{L}$ represents the cross-entropy loss; $\textbf{y}_{slot}$ and $\textbf{y}_{weak}$ represent one-hot vectors of the slot label and the weak label of the sample respectively; the hyperparameter $\alpha$ adjusts how much weak supervision loss contributes to the final loss.

\subsection{Uncertainty-based Criteria}
\label{uncertainty}
In this section, we introduce three commonly-used uncertainty-based active learning strategies. We test the performance of each and integrate them into the proposed Bi-criteria active learning scheme to find the best setting. \\

\noindent\textbf{Entropy Sampling:} 
 Given the predicted probability distribution $\hat{\textbf{y}}_{slot}$, the entropy score will be:
\begin{equation}
C_{entropy} = -\sum_{i} \hat{y}^{(i)}_{slot} log(\hat{y}^{(i)}_{slot}),
\end{equation}
where $i$ denotes each dimension of these vectors. This strategy selects samples where $C_{margin} \ge \tau_e$, where $\tau_e$
is a hyperparameter.\\

\noindent\textbf{Margin Sampling:}  Margin score is defined as the difference between the highest probability $\overline{y}_{slot}$ and the second highest probability $\tilde{y}_{slot}$ obtained from the predicted distribution $\hat{\textbf{y}}_{slot}$, \textit{i.e.}:
\begin{equation}
C_{margin} = \overline{y}_{slot} - \tilde{y}_{slot}.
\end{equation}
This strategy tries to find hard samples where $C_{margin} \le \tau_m$, where $\tau_m$
is a hyperparameter. \\

\noindent\textbf{Bayesian Active Learning by Disagreement (BALD):} As discussed in \cite{srivastava2014dropout}, models with activating dropout produce a different output during multiple inferences. BALD \cite{houlsby2011bayesian} computes model uncertainty by exploiting the variance among different dropout results. Suppose $\overline{y}_{slot}^t$ is the best scoring output for $X$ in the $t-{th}$ forward pass and $T$ is the number of forward passes with a fixed dropout rate, and then we have:
\begin{equation}
C_{BALD} = 1 -\frac{ count(mode(\overline{y}_{slot}^1,\overline{y}_{slot}^2,\cdots,\overline{y}_{slot}^T))}{T},
\end{equation}
where the $mode(\cdot)$ operation finds the output which
is repeated most times, and the $count(\cdot)$ operation counts the number of times this output was repeated. This strategy selects unlabeled samples with $C_{BALD} \ge \tau_b$, where $\tau_b$ is a hyperparameter.\\

%\noindent\textbf{Active$^2$ Learning:} As discussed in \cite{hazra2021active2}, \\

\subsection{Infusing Diversity}
Simply relying on uncertainty-based criteria would invite the redundancy problem where samples of similar semantics and context are selected. To address this, we infuse diversity into the sampling strategy. Inspired by Maximal Marginal Relevance (MMR) in Information Retrieval \cite{carbonell1998use}, we develop a Bi-criteria sampling method which selects those unlabeled samples with high uncertainty and also diverse in meaning at the same time. If we adopt the margin score as the uncertainty score, then the Bi-criteria score for each unlabeled sample $q$ should be:
\begin{equation}
%C_{bi-criteria} = \beta   C_{margin}(s) - (1-\beta)  \max_{s_k \in \mathcal{K}} \textit{Sim}(\textbf{r}_k, \textbf{r}_s),\label{mmr}
C_{bi-criteria} = \beta   C_{margin}(q) - (1-\beta)  \max_{p \in \mathcal{P}} \textit{Sim}(\textbf{r}_l^p, \textbf{r}_u^q),\label{mmr}
\end{equation}

where $\mathcal{P}$ is the set of all labeled samples and $p$ is the index of the labeled sample; $\textbf{r}$ is the vector representation of the sample obtained from Equation \eqref{get_r}; \textit{Sim} stands for the cosine similarity between two representation vectors; $\beta$ is the hyperparameter that controls the contribution of uncertainty and diversity. Specially, when $\beta$ is set to 0, we get the purely diversity-based score as:
\begin{equation}
%C_{diversity}' = -\max_{s_k \in \mathcal{K}} \textit{Sim}(r_k, r_u).
C_{diversity}' = -\max_{p\in \mathcal{P}} \textit{Sim}(r_l^p, r_u^q).
\end{equation}
Intuitively, the Diversity Sampling selects unlabeled samples by their distances from the nearest labeled sample in the feature space. The larger that distance is, the more different in meaning the sample is from labeled sample sets.

On the other hand, when $\beta$ is set to 1, Bi-criteria will be reduced to Margin Sampling, where diversity is no longer taken into account.

\section{Experiments}
\subsection{Datasets}
We follow the datasets listed in \cite{hudevcek2021discovering}. However, the number of slots in the CamRest and
Cambridge SLU datasets is relatively limited considering their dataset size. In our active learning setting, a random portion of initial data is needed to start the training. The initial sets of these two datasets often cover all slots, so we ignore these two datasets with limited slots and mainly conduct experiments on the three large-scale datasets from different domains: \textbf{ATIS} \cite{hemphill1990atis} is a widely used dialogue corpus in flights domain; \textbf{WOZ-attr} \cite{eric2020multiwoz} and \textbf{WOZ-hotel} \cite{eric2020multiwoz} are selected from a large-scale dataset MultiWOZ with attraction domain and hotel domain, respectively. The statistic of the three datasets is shown in Table.~\ref{dataset}. The number of the 
known slots is obtain based on the initial randomly labeled data by 5\%.

%	\begin{tabular}{p{1.5cm}|c|c|c}
%\vspace{-0.2cm}
\begin{table}[htb]
\small
\caption{The statistic information of three datasets}
%\vspace{-0.2cm}
\begin{tabular}{p{1.5cm}cccccc}
\hline
\xrowht[()]{4pt}
\multirow{2}{*}{Dataset} & \multirow{2}{*}{Domain} & \multirow{2}{*}{\#Samples} & \multicolumn{3}{c}{\#Slots} \\ \cline{4-6} 
                         &                    &        & Known       & New       & Total  \\ \hline \xrowht[()]{3pt}
ATIS                     & Flight               & 4,978               & 54         & 25   & 79       \\  \xrowht[()]{5pt}
WOZ-attr                 & Attraction            & 7,524              & 4         & 4     & 8     \\  \xrowht[()]{5pt}
WOZ-hotel               & Hotel                  & 14,435          & 4       & 5     & 9    \\ \hline \xrowht[()]{5pt}           
\end{tabular}
\label{dataset}
\vspace{-0.4cm}
\end{table}

\vspace{-0.2cm}
\subsection{Implementation Details}
We apply the pre-trained ‘bert-base-cased' version of BERT \cite{kenton2019bert} to implement our model. We adopt Adam strategy \cite{DBLP:journals/corr/KingmaB14} for optimization with the base
learning rate of 5e-5. The linear decay of the learning rate is applied following \cite{kenton2019bert}. The number of max initial training epochs is 30 and the batch size is 128. For each follow-up active learning iteration, we fine-tune the model on the updated labeled training set for two epochs following \cite{hazra2021active2}.

We split each dataset into \textit{training / testing / validation} sets $(0.8/0.1/0.1)$. Each of our experiments is an emulation of the active learning cycle: selected instances are not presented to experts for annotation but are labeled automatically according to the gold standard. A random 5\% of the whole training set is chosen as a warm-up dataset by the random seed 0. At each active learning iteration, 2\% of new training samples are selected for annotation. For selection strategies based on the Monte Carlo dropout, we make five stochastic predictions.

Since not all the slot labels are discovered during each iteration, we apply a label mask during loss calculation and active learning sampling. Like the operation in Equation \ref{mask}, before calculating the criteria scores in Subsection \ref{uncertainty}, the predicted probability distribution is also multiplied by a mask vector to make sure undiscovered slot labels are invisible to the sampling process.

\subsection{Evaluation Metric}
We evaluate the performance via the widely used classification metric, \textit{i.e.}, F1-score \cite{sang2003introduction}. Suppose the ground-truth slot values are  $\mathcal{M}_{1}, \mathcal{M}_{2}, ..., \mathcal{M}_{n}$, where $n$ denotes the number of slots. The predicted values are $\mathcal{E}_{1}, \mathcal{E}_{2}, ..., \mathcal{E}_{n}$. Then for each slot type $i$, the precision and recall score are calculated as:

\begin{align}
    P_{i} = &\frac{|\mathcal{M}_{i}\cap\mathcal{E}_{i}|}{|\mathcal{E}_{i}|}, \\
    R_{i} = &\frac{|\mathcal{M}_{i}\cap\mathcal{E}_{i}|}{|\mathcal{M}_{i}|}.
\end{align}

Then weighted overall precision and recall score $P$ and $R$ are calculated as:

\begin{align}
    P = \sum^n_{i=1}&\frac{|\mathcal{E}_{i}|}{\sum^n_{j=1}|\mathcal{E}_{j}|}P_{i}, \\
    R = \sum^n_{i=1}&\frac{|\mathcal{M}_{i}|}{\sum^n_{j=1}|\mathcal{M}_{j}|}R_{i}.
\end{align}

Finally, we obtain the F1 score as:

\begin{align}
    F1 = &\frac{2PR}{P+R}.
\end{align}

Since the F1 score is calculated based on slot value spans, it is also called Span-F1 in this paper.

\subsection{Competitive Methods}
We compare our method with two groups of baselines: active learning methods and semi-supervised methods with 21\% randomly labeled data. We utilize the same backbone for different methods for fair comparison.
%Active learning strategies for comparison can be classified as:

\begin{itemize}
    \item \textbf{Active learning methods}
    \begin{itemize}
        \item \textbf{Random}: The active learning method with random sampling strategy.
        \item \textbf{Uncertainty-based sampling}:  We compare our methods with several uncertainty-based strategies mentioned before including \textbf{Entropy} , \textbf{Margin} and \textbf{BALD} sampling.
        \item \textbf{Diversity-based sampling}: As mentioned before, we set $\beta$ as 0 to achieve the pure diversity sampling method. 
        \item \textbf{Hybrid} sampling: 
        Active$^2$ Learning \cite{hazra2021active2} is a two-stage hybrid sampling method. It first utilizes an uncertainty-based criterion to select a coarse sample set. Then an external corpus is adopted to assist the clustering step in order to ensure the diversity. To adapt \cite{hazra2021active2} for a fair comparison, we choose the Margin Sampling as the uncertainty-based criterion. Then we naturally apply the weak labels obtained from the language tools to replace the extra clustering step in the second stage.       
    \end{itemize}
    \item \textbf{Semi-supervised methods}: As we formulate the new slot discovery in an IE fashion, we actually transform the problem into an instance (one value candidate and its context) class discovery task which is rather close to the intent discovery setting. Hence, we compare with the state-of-arts semi-supervised intent discovery methods \textbf{CDAC+} \cite{lin2020discovering} and \textbf{DeepAligned} \cite{zhang2021discovering}. We adapt them to our new slots discovery task since they are designed as a classification scheme.
\end{itemize}

\subsection{Quantitative Results}

\subsubsection{Active \textit{v.s.} Semi-supervised} 
We report the results compared with semi-supervised methods in Table~\ref{semi}. We can observe that our method outperforms all the semi-supervised methods on all three datasets. The proposed method surpasses the second-best method on ATIS, WOZ-attr, WOZ-hotel by 24.66\%, 11.53\%, and 25.08\% respectively. The result demonstrates the effectiveness of using active learning and the strength of human labeling efforts.

It is also shown that the DeepAligned method has better performance than CDAC+. Specifically, DeepAligned outperforms CDAC+ by 3.23\%, 8.72\%, 27.35\% respectively on ATIS, WOZ-attr, and WOZ-hotel. It is worth noticing that there is a huge performance drop for the two methods on WOZ-hotel dataset. We suspect it is attributed to the fact that the distribution of the WOZ-hotel dataset is difficult to fit, especially for the CDAC+ method which overemphasizes the pairwise similarity as prior knowledge.
%\vspace{-0.2cm}
\begin{table}[!htp]
	\small
	\caption{Comparison with other competitive semi-supervised methods. Here we provide the Span-F1 score.}
%	\vspace{-0.2cm}
	\label{semi}
	\renewcommand*{\arraystretch}{1.3}
	\begin{center}
		\scalebox{1}{
			\begin{tabular}{p{2.5cm}|c|c|c}
				\hline
				\xrowht[()]{5pt}
				\multirow{1}{*}{\textbf{Method}} 
				& \multicolumn{1}{c|}{\textbf{ATIS}} &  \multicolumn{1}{c|}{\textbf{WOZ-attr}} &  \multicolumn{1}{c}{\textbf{WOZ-hotel}}\\
				\hline
				%\textit{BERT-DTC}      & 58.65 & 60.88 & 19.32       \\\xrowht[()]{3pt}
				\textit{CDAC+} \cite{lin2020discovering}         & 60.07 & 58.00   & 16.51     \\\xrowht[()]{3pt}
				\textit{DeepAligned} \cite{zhang2021discovering} & 63.30   & 66.72  & 43.86    \\%\xrowht[()]{3pt}
				Ours \textit{(Bi-Criteira)} & \textbf{87.96}   & \textbf{78.25}  & \textbf{68.94}  \\
				\hline
			\end{tabular}
		}
	\end{center}
%\vspace{-0.3cm}
\end{table}

\subsubsection{Bi-Criteria \textit{v.s.} Other Active Strategies} \
\begin{figure*}[!htp]
	\centering
	\subfigure[ATIS]{\includegraphics[width=0.32\linewidth]{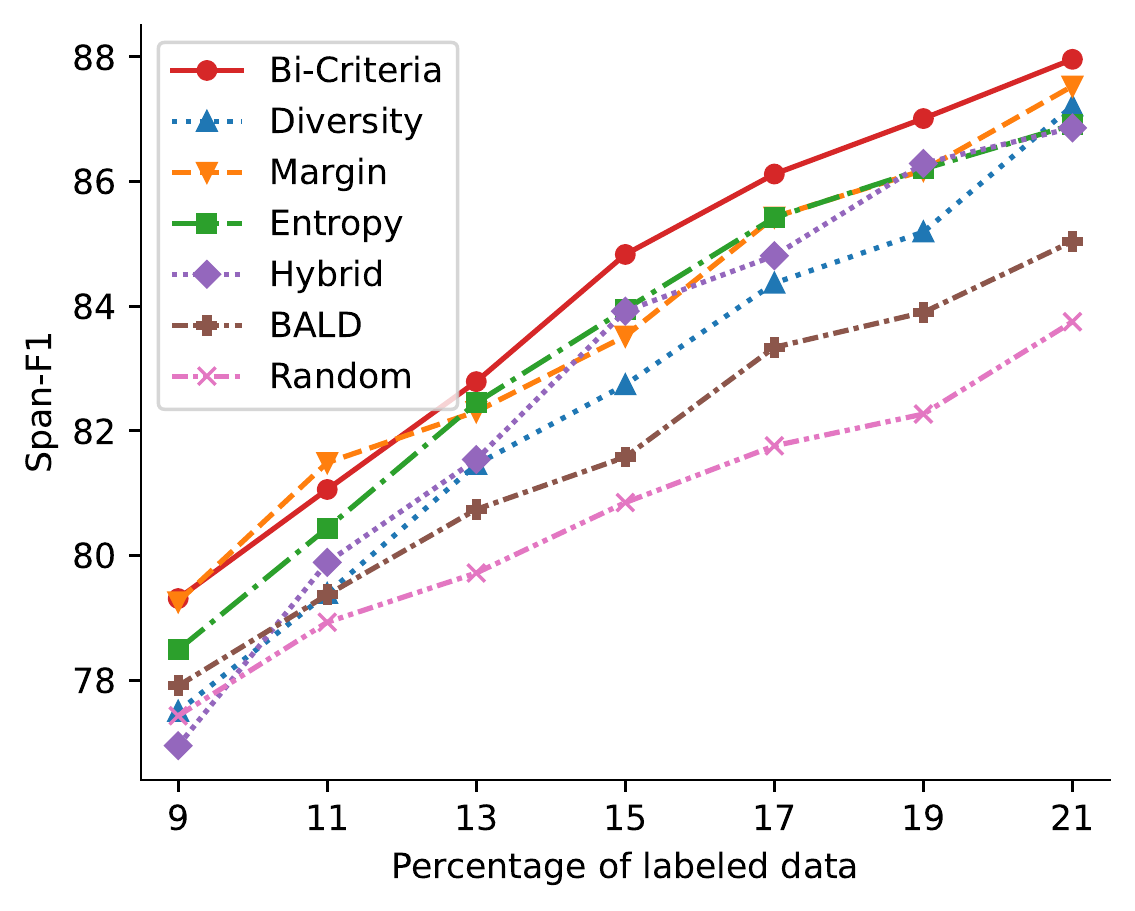}}
	\subfigure[WOZ-attr]{\includegraphics[width=0.32\linewidth]{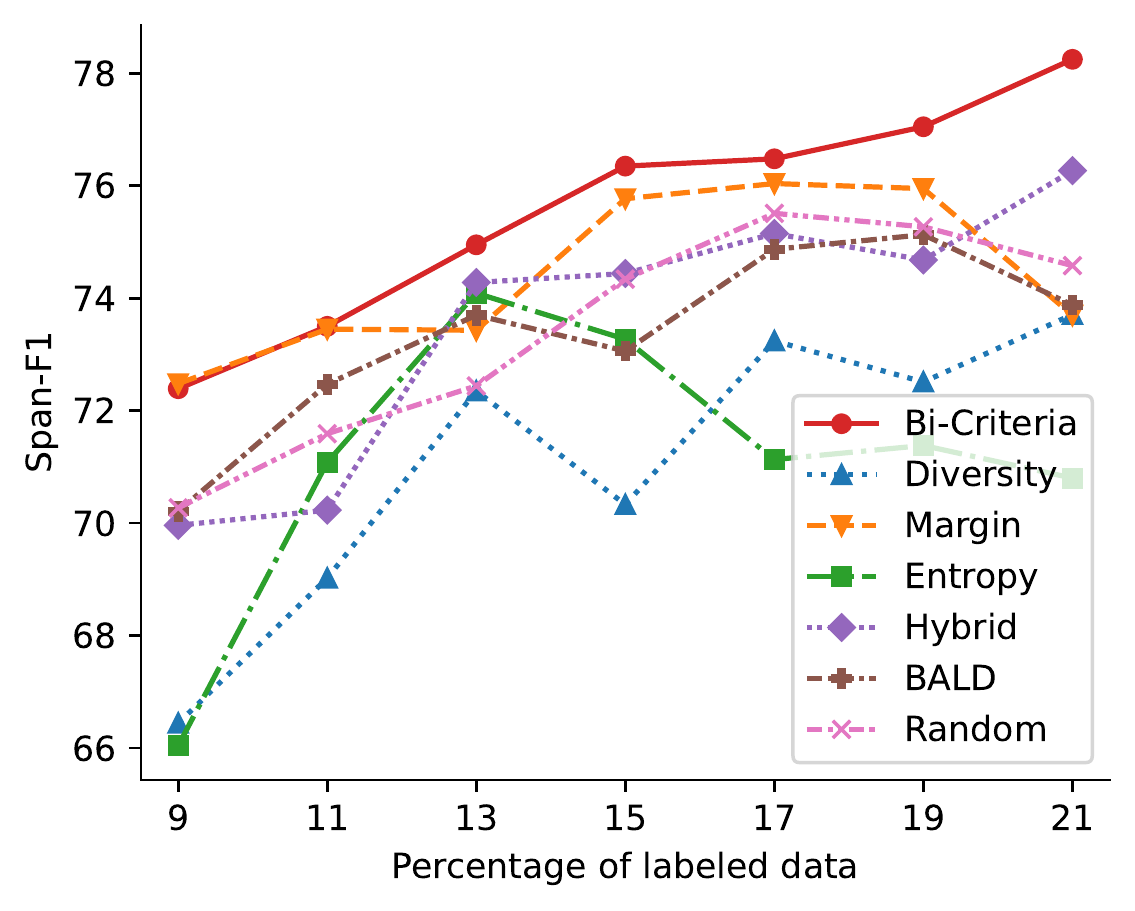}}
	\subfigure[WOZ-hotel]{\includegraphics[width=0.32\linewidth]{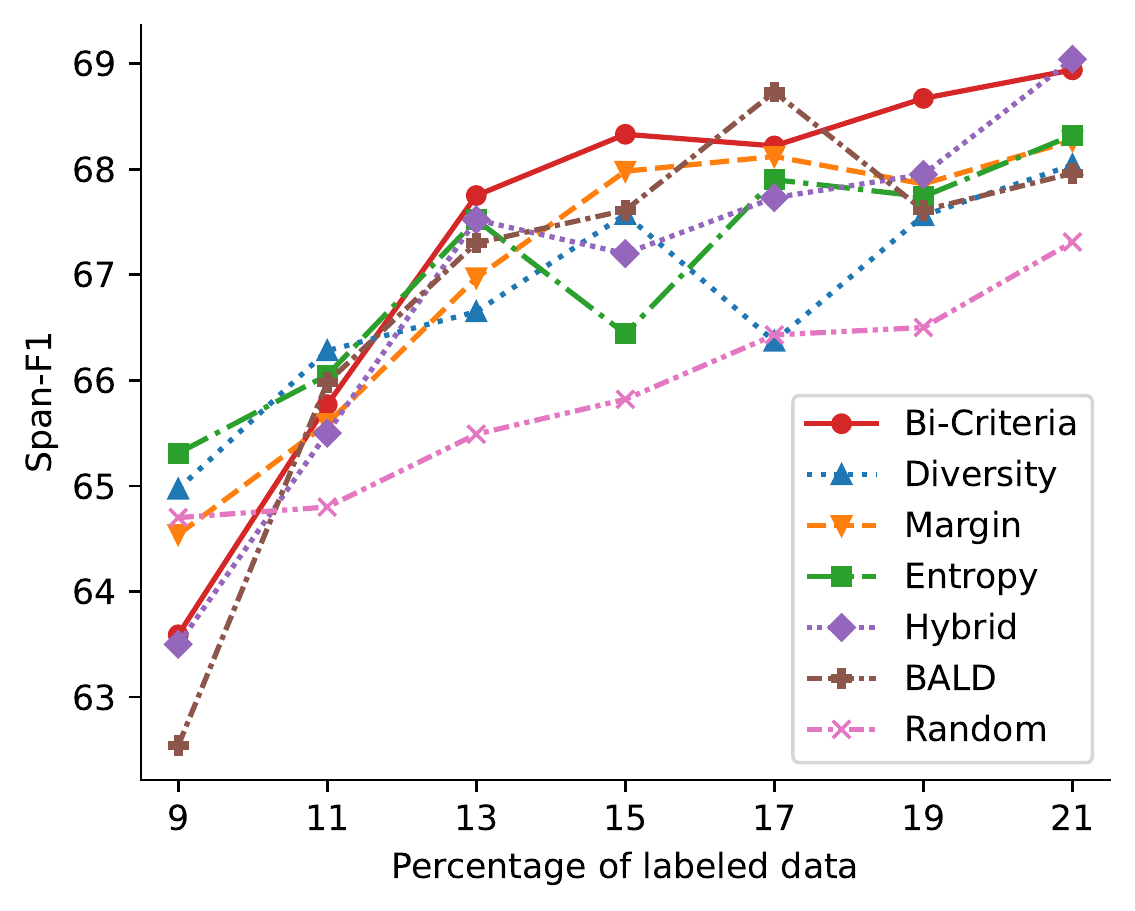}}
	%\vspace{-0.3cm}
	\caption{The results of different active learning strategies on the three public datasets. All methods start from the same initial training checkpoint over 5\% randomly sampled instances. These plots have been magnified to highlight the regions of interest.}
	%\vspace{-0.2cm}
	\label{strategy-comp}
\end{figure*}
The results of experiments on three public datasets with different active learning strategies are presented in Figure \ref{strategy-comp}. Due to the intrinsic discrepancy among datasets, we set the $\alpha$ in Equation \eqref{final_loss} for each dataset differently (0.05 on ATIS and WOZ-attr, 0.1 on WOZ-hotel).
%Although the $\alpha$ in Equation \eqref{final_loss} that yields the best results is conditioned by specific dataset (0.05 on ATIS and WOZ-attr, 0.1 on WOZ-hotel), due to the intrinsic discrepancy among datasets, it remains unchanged during experiments on each dataset.
As seen, the F1 scores significantly vary among different active learning strategies, and Bi-criteria generally performs the best on all three datasets in terms of accuracy and stability. The mean of differences between the best score of bi-criteria and the best scores among other sampling strategies over all sampling steps is 0.61\% on ATIS and 0.95\% on WOZ-attr. On WOZ-hotel, though surpassed by BALD and Hybrid strategy at the 17 and 21 percent stages, Bi-criteria exhibits performance with less fluctuation thus better stability.

As expected, Random Sampling strategy is generally overwhelmed by most active learning strategies most of the time, since neither redundancy nor diversity is concerned during the data selection. However, this tendency is less conspicuous on WOZ-attr, where Entropy Sampling and BALD perform worst. 

Note that Margin Sampling and Diversity Sampling are the special cases of the Bi-criteria strategy when $\beta$ in Equation \eqref{mmr} is set to $\beta=1$ and $\beta=0$ respectively. It is easily observed from Figure \ref{strategy-comp} that Bi-criteria strategy outperforms both of the strategies in Span-F1 and stability. As the mixture of Margin Sampling and Diversity Sampling, Bi-criteria takes advantage of both uncertainty and diversity. It indicates that these two strategies are both essential components in terms of active selection and impact the results in a cooperating way to some extent. Further analysis could be found in Subsection \ref{beta}.

%\vspace{-0.1cm}
\subsection{Ablation Studies and Further Analysis}

\subsubsection{Effect of hyperparameter $\alpha$}
\begin{figure*}[!htp]
	\centering
	\subfigure[ATIS]{\includegraphics[width=0.32\linewidth]{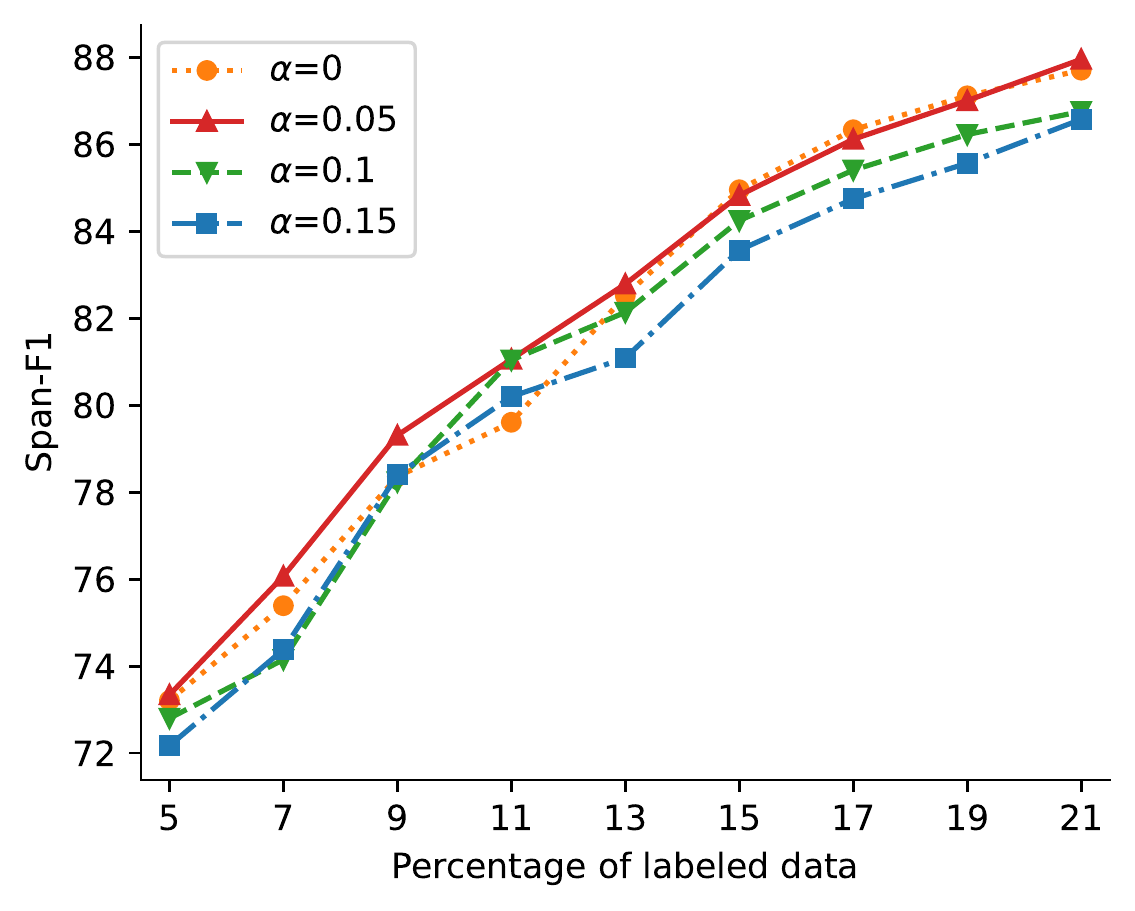}}
	\subfigure[WOZ-attr]{\includegraphics[width=0.32\linewidth]{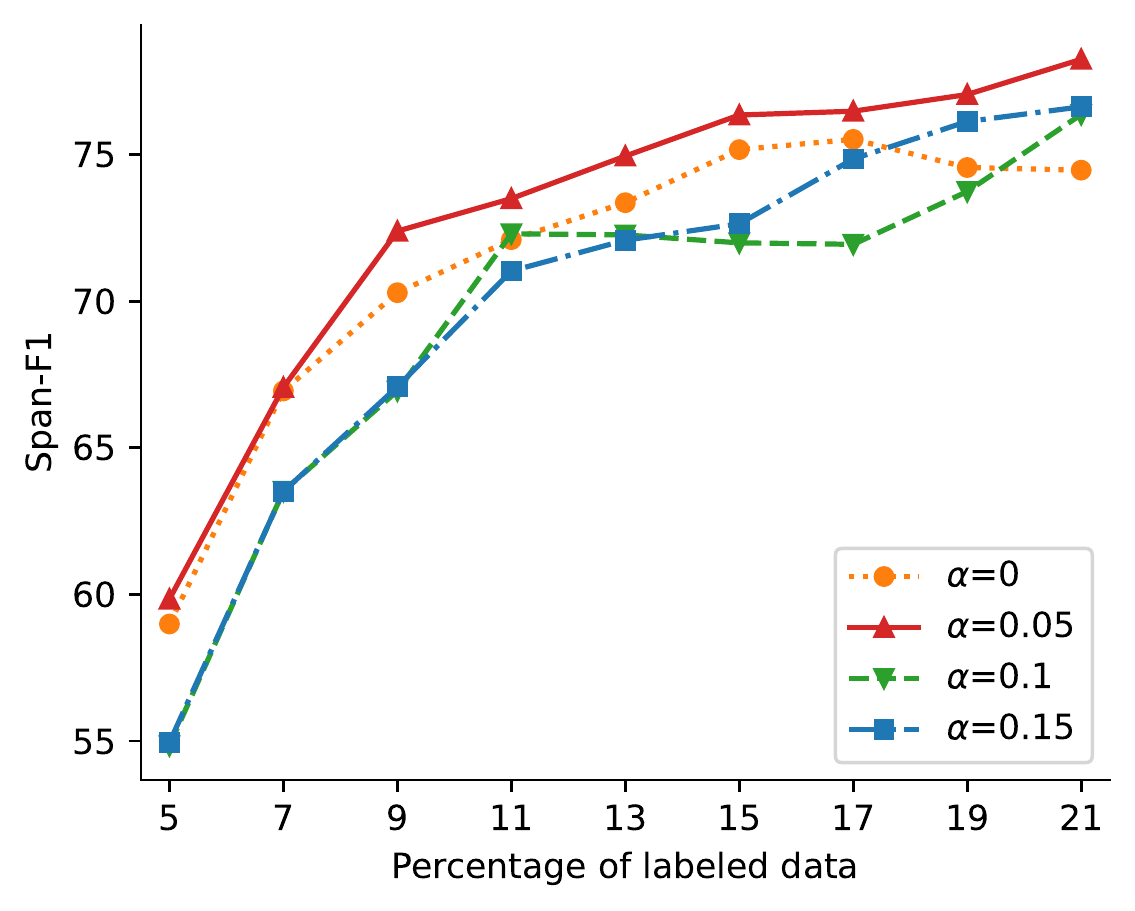}}
	\subfigure[WOZ-hotel]{\includegraphics[width=0.32\linewidth]{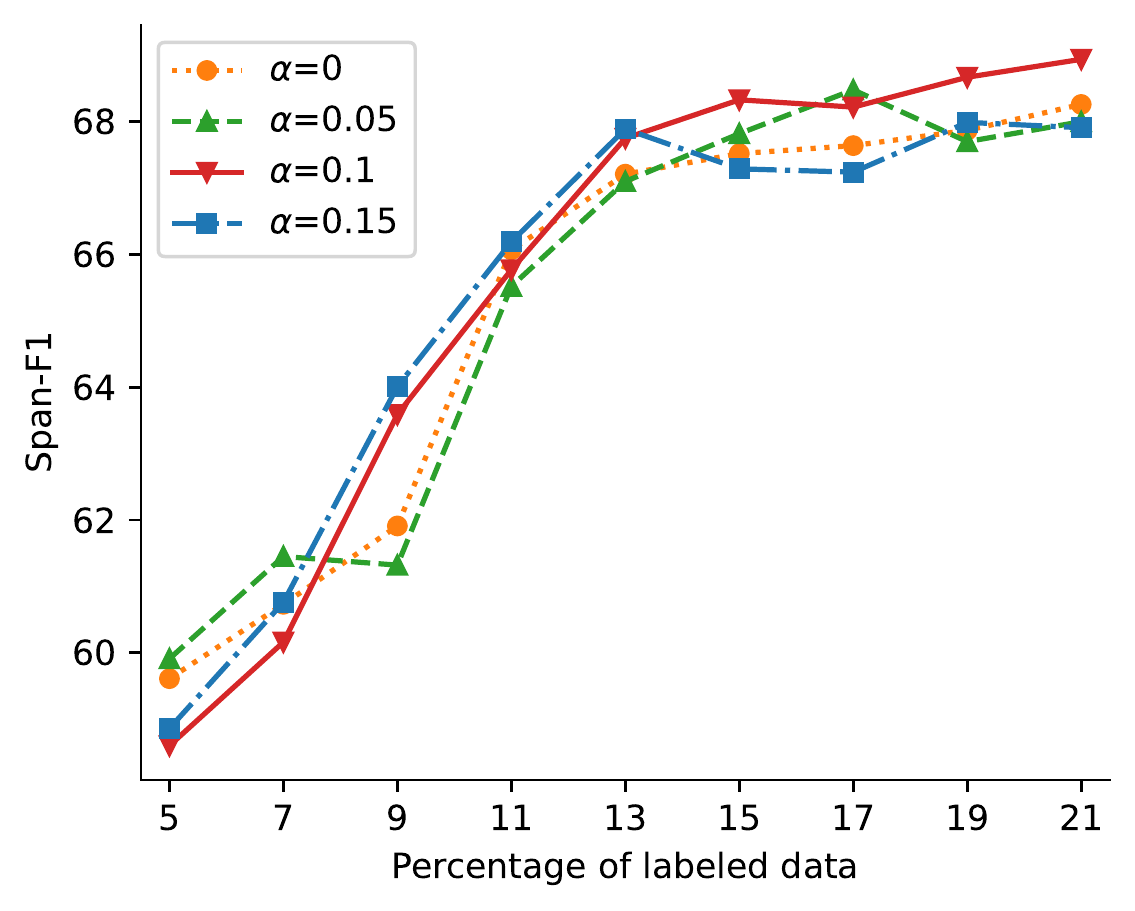}}
%	\vspace{-0.4cm}
	\caption{Ablation study of $\alpha$ on three public datasets. All methods start from the same initial training checkpoint over 5\% randomly sampled instances. These plots have been magnified to highlight the regions of interest.}
	\label{alpha-comp}
\end{figure*}
We fix the $\beta$ in Equation \eqref{mmr} and adjust $\alpha$ in Equation \eqref{final_loss} to see its effect on the performance of Bi-criteria active learning strategy. The hyperparameter $\alpha$ indicates the proportion of weak supervision loss in the final loss. The higher $\alpha$ means the greater contribution of weak supervision to the result. Specially, the model does not learn any semantics from weak supervision when $\alpha$ is set to 0. According to our observation, the $\alpha$ tends to have a relatively small effect on the performance compared with other parameters and therefore only four value settings are tested and shown here in Figure \ref{alpha-comp}.

As is seen from the line charts in Figure \ref{alpha-comp}, tuning $\alpha$ to 0.05 leads to the performance with both better Span-F1 and stability compared with other $\alpha$ settings on ATIS and WOZ-attr while $\alpha$ at 0.1 results in the best stability and relatively high Span-F1 on WOZ-hotel. Moreover, method with $\alpha$ at $0$ does not perform best on all three datasets, which validates the usefulness of weak supervision. The mean of differences between the Span-F1 of the selected $\alpha$ (red line in the graphs) and the Span-F1 of $\alpha$ at 0 over all active learning steps is 0.36\%, 1.61\%, 0.36\% on ATIS, WOZ-attr, and WOZ-hotel respectively. This result proves that weak supervision indeed boosts the performance of model on our task, thus necessitating the adoption of our multi-task network structure.

However, the performance does not necessarily improve as the proportion of weak supervision grows higher. This tendency is easily observed from the results on ATIS, where the performance declines as $\alpha$ grows bigger from 0.05. Therefore, finding an appropriate weight for weak supervision is critical to our multi-task network.

%\vspace{-0.1cm}
\subsubsection{Effect of hyperparameter $\beta$}
\label{beta}
\begin{figure*}[!htp]
	\centering
	\subfigure[ATIS]{\includegraphics[width=0.32\linewidth]{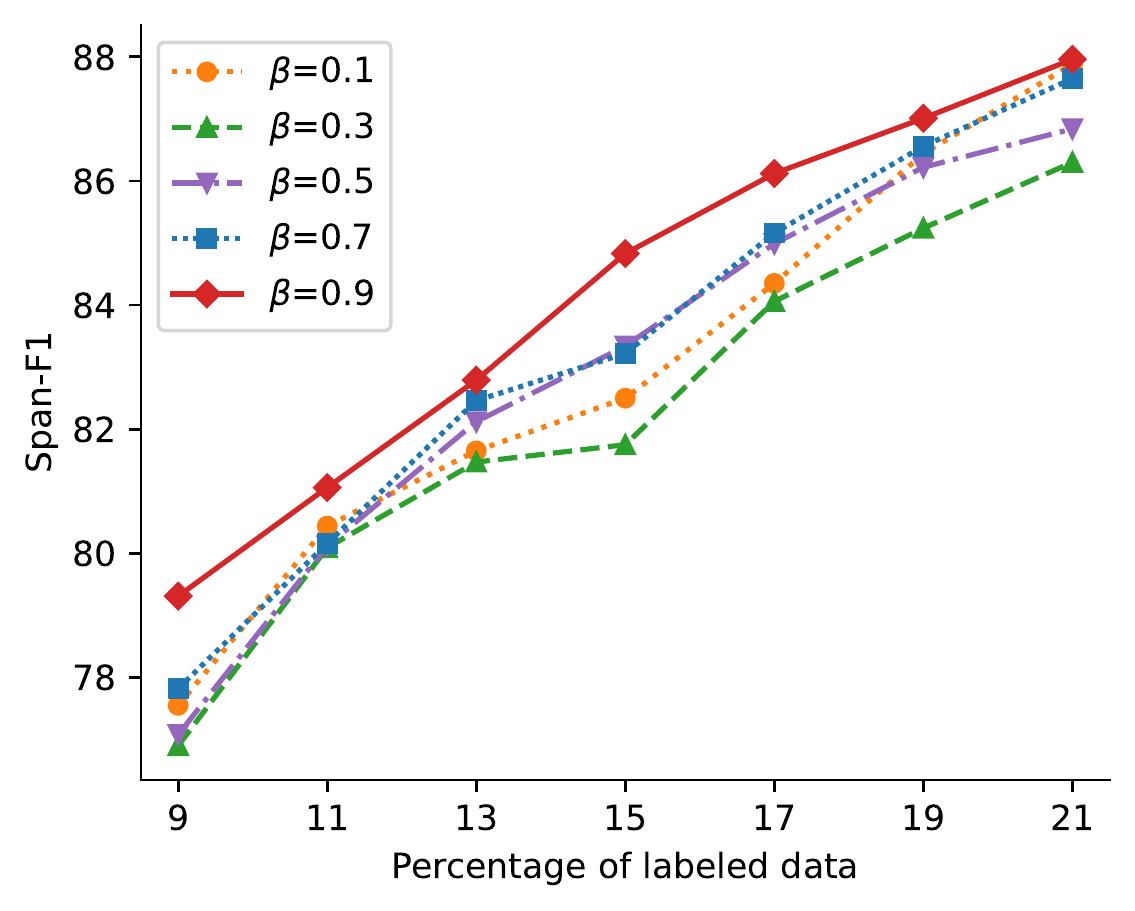}}
	\subfigure[WOZ-attr]{\includegraphics[width=0.32\linewidth]{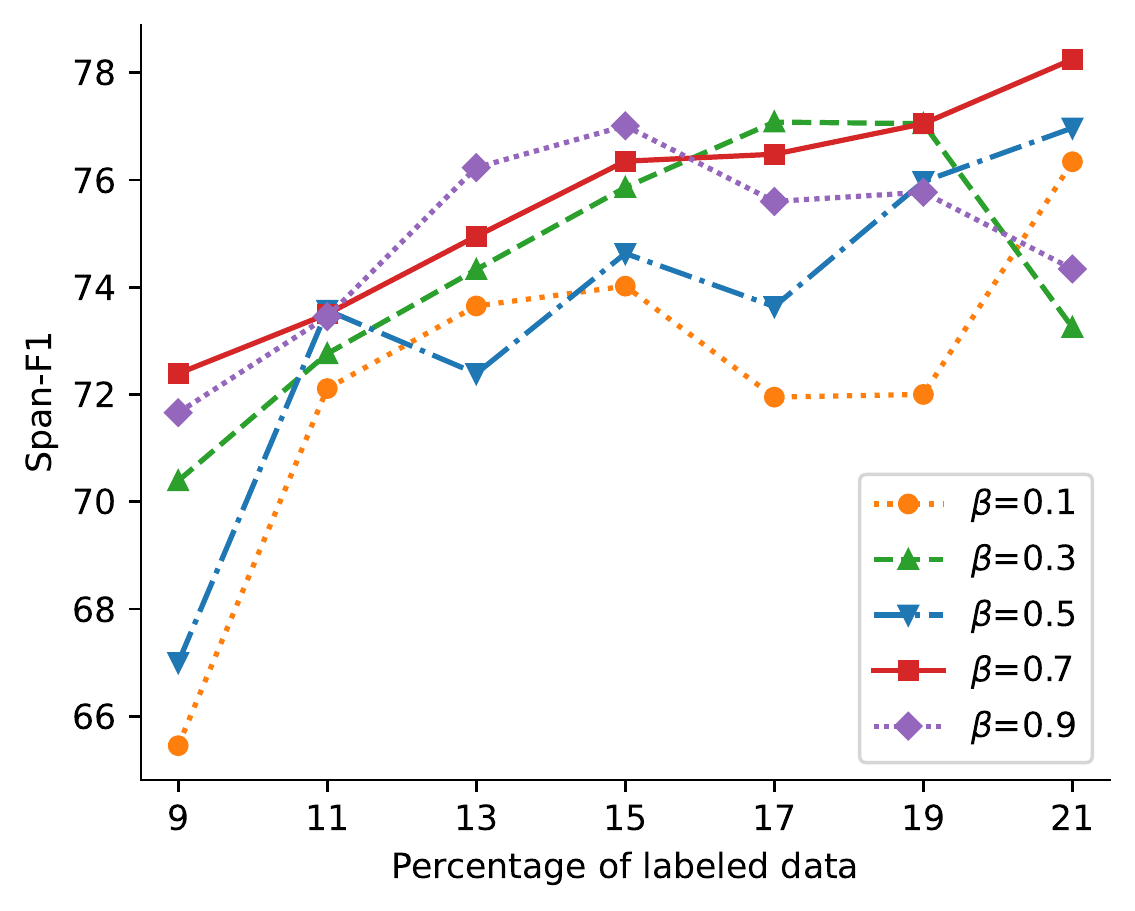}}
	\subfigure[WOZ-hotel]{\includegraphics[width=0.32\linewidth]{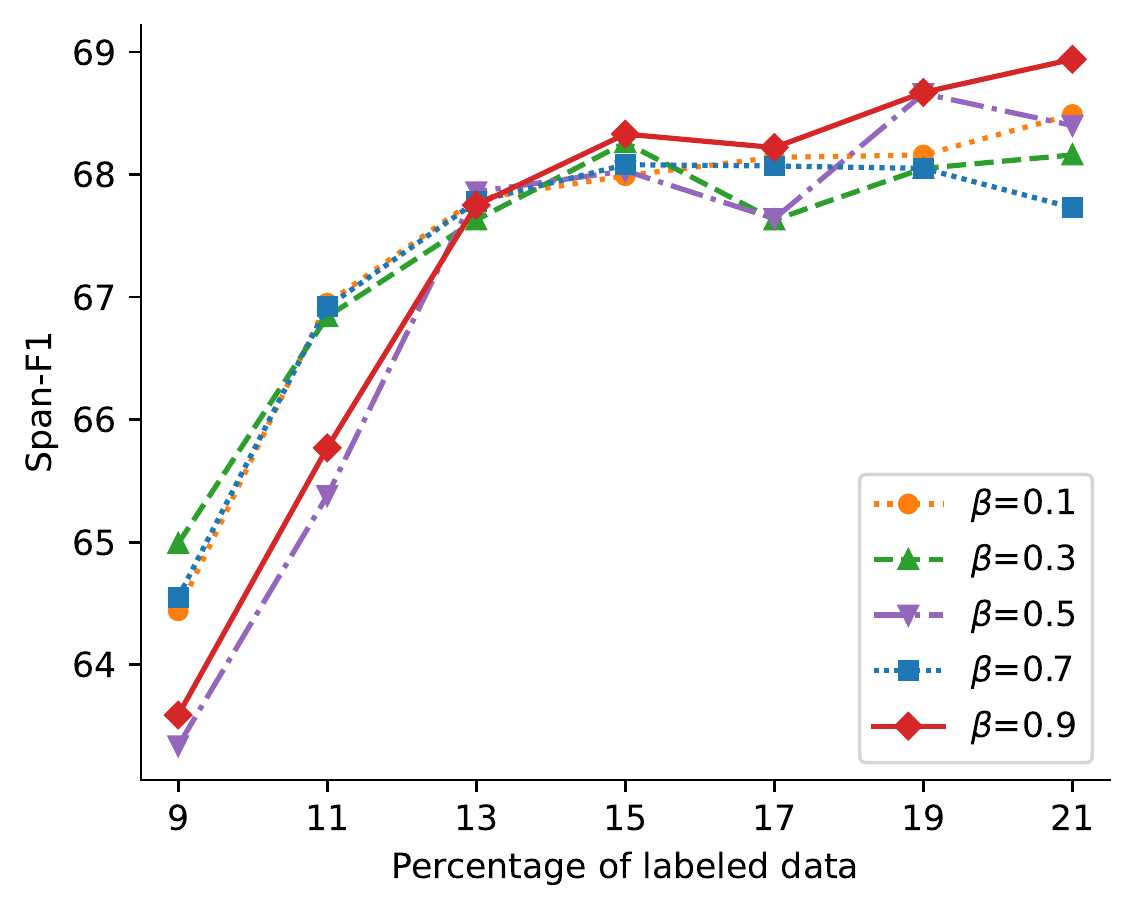}}
%	\vspace{-0.4cm}
	\caption{Ablation study of $\beta$ on the three public datasets. All methods start from the same initial training checkpoint over 5\% randomly sampled instances. These plots have been magnified to highlight the regions of interest.}
	\label{beta-comp}
\end{figure*}

We also study the effect of the $\beta$ in Equation \eqref{mmr}. Note that $\beta=0$ and $\beta=1$ are equivalent to purely Diversity Sampling and Margin Sampling respectively, performances of which have been shown in Figure \ref{strategy-comp}. 
In general, the Bi-criteria method incorporating both uncertainty-based and diversity-based strategies tends to yield better results compared to using either strategy alone. Moreover, the weights of these two aspects also exert certain influence on the performance. As Figure \ref{beta-comp} shows, the Bi-criteria method achieves satisfying results when $\beta$ is set to 0.9 on ATIS and WOZ-hotel and 0.7 on WOZ-attr. Experiments with $\beta$ below or equal to 0.5 generally achieve poor results compared to settings with higher $\beta$. This indicates that uncertainty actually contributes more to the overall performance. However, the diversity signal is still indispensable since it helps to achieve results that Margin Sampling itself cannot.

%\vspace{-0.1cm}
\subsubsection{Comparison with different kinds of weak supervision} 
\label{pretrain}
We also explore different ways of using weak supervision. The key goal of weak supervision is to make use of existing weak labels to facilitate our task. In our proposed method, weak supervision is implemented in a multi-task fashion. Labels given by language tools are adopted to conduct an individual classification task, whose loss contributes to the final loss. The alternative way is to pre-train the BERT model with these labels for classification first, and then fine-tune the BERT parameters for the new training phase for our new slot discovery task with new classifier head.

\begin{table}[!htp]
	\small
	\caption{Comparison with different kinds of weak supervision on three datasets. Here we provide the Span-F1 score.}
	%\vspace{-0.3cm}
	\label{weak-supervision}
	\renewcommand*{\arraystretch}{1.2}
	\begin{center}
		\scalebox{1}{
			\begin{tabular}{p{1.4cm}|cc|cc|cc}
				\hline
				\xrowht[()]{3pt}
				\multirow{2}{*}{\textbf{Method}} 
				& \multicolumn{2}{c|}{\textbf{ATIS}} &  \multicolumn{2}{c|}{\textbf{WOZ-attr}} &  \multicolumn{2}{c}{\textbf{WOZ-hotel}}\\
				\cline{2-7}
				\xrowht[()]{4pt}
				&Start&End&Start&End&Start&End\\
				\hline\xrowht[()]{5pt}
				\textit{no weak.} & 73.21 & 87.71 & 58.14 & 75.38 & 59.61 & 68.26 \\\xrowht[()]{3pt}
				\textit{pretrain} & \textbf{74.61} & 87.85 & 59.21 & 74.15 & \textbf{61.86} & 67.95 \\\xrowht[()]{3pt}
				\textit{multi-task} & 73.34 & \textbf{87.96} & \textbf{59.84} & \textbf{78.25} & 58.60 & \textbf{68.94} \\
				\hline
			\end{tabular}
		}
	\end{center}
\vspace{-0.4cm}
\end{table}

Table \ref{weak-supervision} shows the results under different kinds of weak supervision. These results represent the Span-F1 at the start point (5\% labeled data) and the endpoint (21\% labeled data) of the active learning process on three datasets. It can be seen that methods with weak supervision (\textit{pretrain} and \textit{multi-task}) achieve Span-F1 higher than the method without it both at the beginning and the end of the active learning process in all three datasets, which again demonstrates the effectiveness of weak supervision. It is worth noticing that when 5\% training data are labeled, the pretraining method achieves Span-F1 higher than the second best method by 1.27\% and 2.25\% on ATIS and WOZ-hotel respectively. However, when 21\% of training data are labeled, the multi-task method prevails. We can therefore infer that weak supervision as pre-training may enhance the starting point but tend to converge at a lower level than weak supervision as multi-task does in our setting.

\section{Conclusion and Future Work}
In this work, we formulated a general new slot discovery task for task-oriented conversational systems. We designed a bi-criteria active learning scheme for integrating both uncertainty-based and diversity-based active learning strategies. Specifically, to alleviate the limited labeled data problem, we leverage the existing language tools to extract the candidate values and pseudo labels as weak signals. Extensive experiments show that it effectively reduces human labeling effort while ensuring relatively competitive performance. 

We notice that responses to user utterances are abundant in dialogue datasets, which reflects the semantics in user utterances to some extent. Such evidence has the potential to guide the sample selection process in active learning. In the future, we plan to discover new slots by further leveraging such signals. Besides, during the training of AL, we fine-tune the model at each epoch with newly added samples. With the increase of the trained data, the model will encounter the catastrophic forgetting problem. In future work, we will explore a more flexible training strategy to handle this issue.

%\section*{Acknowledgments}
%This should be a simple paragraph before the References to thank those individuals and institutions who have supported your work on this article.


\small
\bibliographystyle{IEEEtranN}
%\balance
\bibliography{anthology}

\vspace{-1cm}
\begin{IEEEbiography}[{\includegraphics[width=1in,height=1.25in,clip,keepaspectratio]{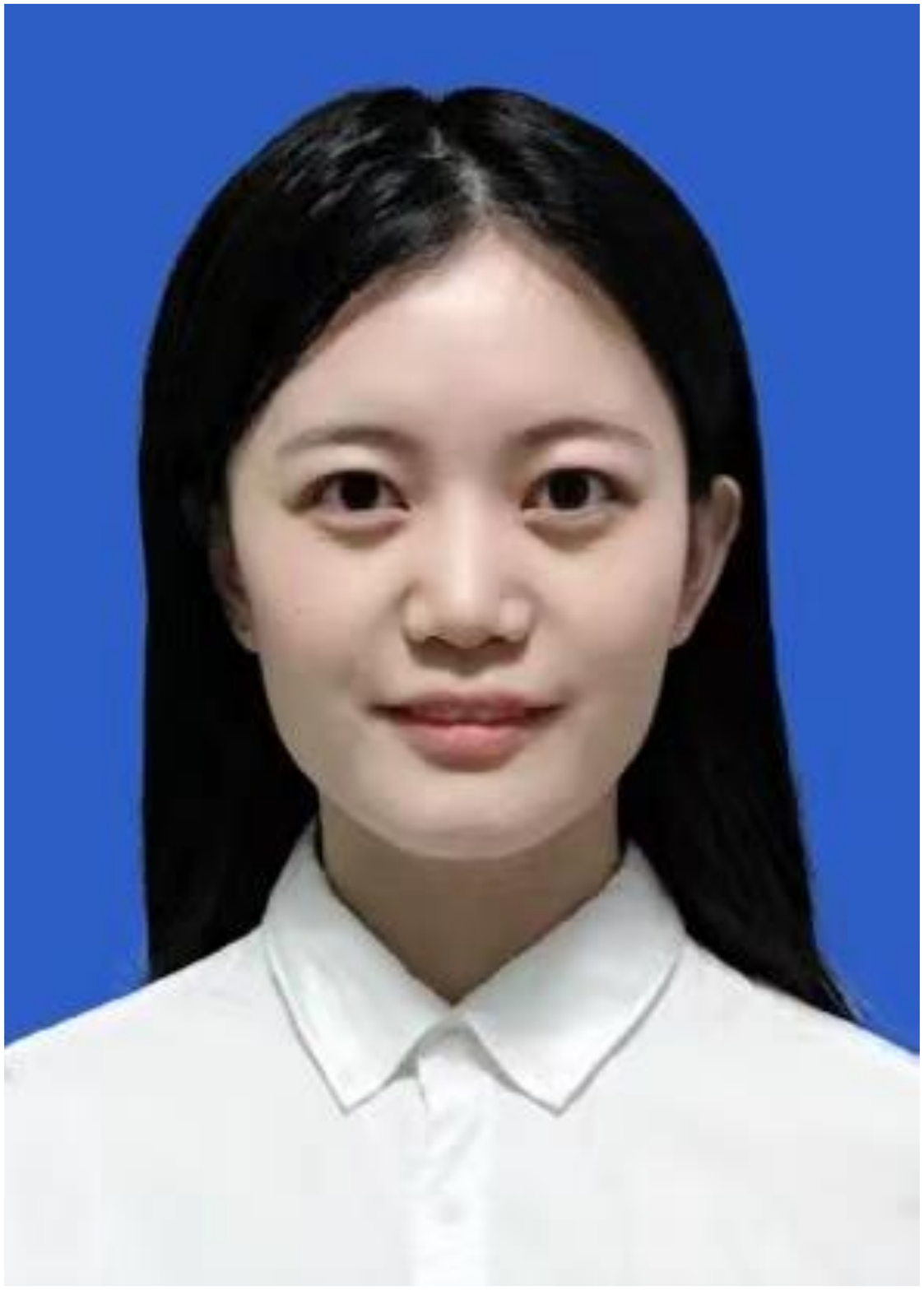}}]{Yuxia Wu}
received the Ph.D. degree from Xi'an Jiaotong University in 2023, the M.S. degree from the Fourth Military Medical University in 2017 and the B.S. degree from Zhengzhou University in 2014. She is currently a research scientist at Singapore Management University. Her research interests include natural language processing, social multimedia mining and recommender systems.
\end{IEEEbiography}

\vspace{-1cm}
\begin{IEEEbiography}[{\includegraphics[width=1in,height=1.25in,clip,keepaspectratio]{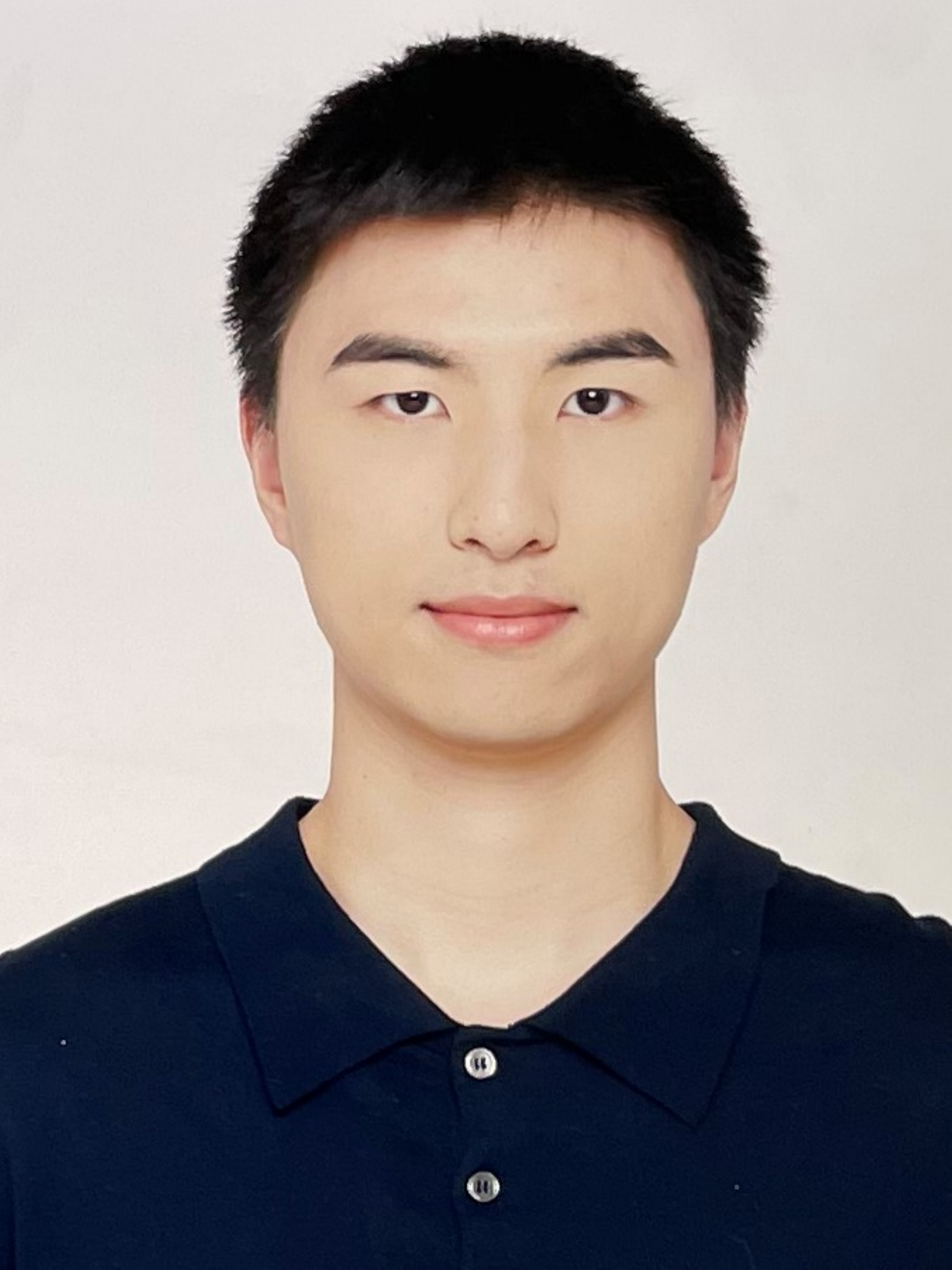}}]{Tianhao Dai} is currently pursuing his undergraduate degree in the School of Cyber Science and Engineering at Wuhan University. From August 2022 to January 2023, he served as a visiting research student at Singapore Management University under the supervision of Assistant Professor Lizi Liao. His research interests include the field of natural language processing and computational linguistics.
\end{IEEEbiography}

\vspace{-1cm}
\begin{IEEEbiography}[{\includegraphics[width=1in,height=1.25in,clip,keepaspectratio]{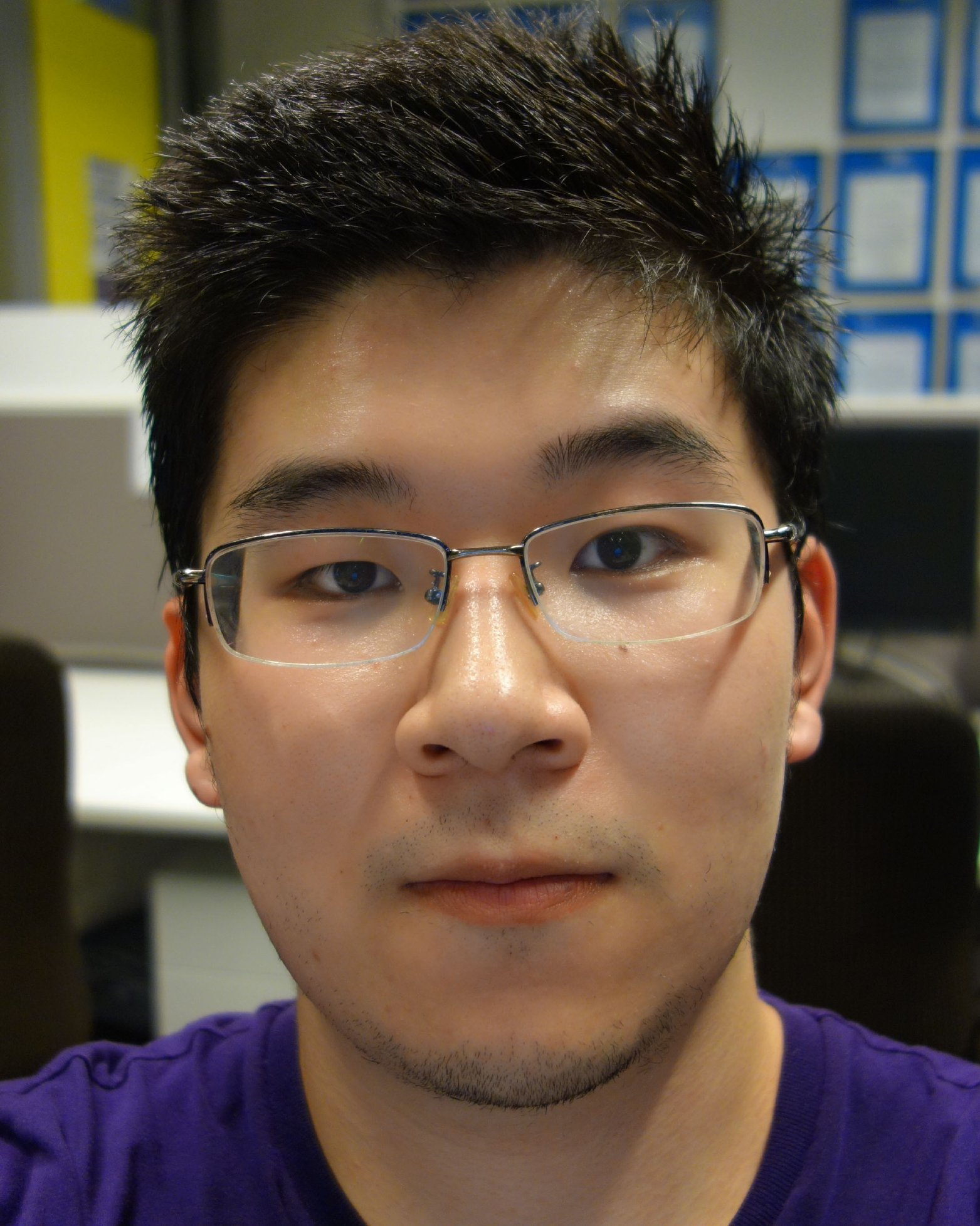}}]{Zhedong Zheng}
received the Ph.D. degree from the University of Technology Sydney, Australia, in 2021 and the B.S. degree from Fudan University, China, in 2016. He is currently a postdoctoral research fellow at the School of Computing, National University of Singapore. He was an intern at Nvidia Research (2018) and Baidu Research (2020). His research interests include robust learning for image retrieval, generative learning for data augmentation, and unsupervised domain adaptation.
\end{IEEEbiography}

\vspace{-1cm}
\begin{IEEEbiography}[{\includegraphics[width=1in,height=1.25in,clip,keepaspectratio]{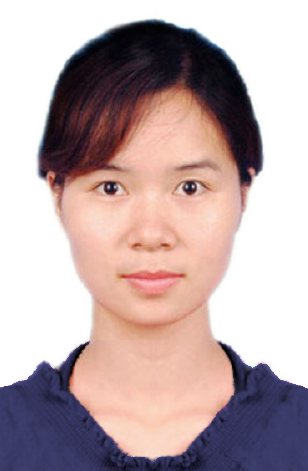}}]{Lizi Liao} is an assistant professor with Singapore Management University. She
received the Ph.D. degree in 2019 from NUS Graduate School for Integrative Sciences and Engineering at
the National University of Singapore. Her research interests include conversational system, multimedia analysis and recommendation. 
Her works have appeared in top-tier conferences such as MM, WWW, ICDE, ACL, IJCAI and AAAI, and top-tier journals such as TKDE. She received the Best Paper Award Honorable Mention of ACM MM 2018. Moreover, she has served as the PC member for international conferences including SIGIR, WSDM, ACL, and the invited reviewer for journals including TKDE, TMM and KBS.
\end{IEEEbiography}

\vfill

\end{document}